\documentclass{article}
\usepackage{amsmath,epsfig}
\usepackage[preprint]{spconf}
\usepackage{cite}

\let\OLDthebibliography\thebibliography
\renewcommand\thebibliography[1]{
  \OLDthebibliography{#1}
  \setlength{\parskip}{0pt}
  \setlength{\itemsep}{0pt plus 0.3ex}
}

\copyrightnotice{\textit{Published in IEEE ICME 2021. Please cite \cite{9428191} while referencing this paper.}}

\begin{document}\sloppy

\def\x{{\mathbf x}}
\def\L{{\cal L}}

\title{ASOC: Adaptive Self-aware Object Co-localization}
%
\name{Koteswar Rao Jerripothula$^{\ast}$, Prerana Mukherjee$^{\dagger}$}
\address{$^{\ast}$CSE Department, Indraprastha Institute of Information Technology Delhi (IIIT-Delhi)\\ $^{\dagger}$School of Engineering, Jawaharlal Nehru University, Delhi}

\maketitle

\begin{abstract}
The primary goal of this paper is to localize objects in a group of semantically similar images jointly, also known as the object co-localization problem. Most related existing works are essentially weakly-supervised, relying prominently on the neighboring images' weak-supervision. Although weak supervision is beneficial, it is not entirely reliable, for the results are quite sensitive to the neighboring images considered. In this paper, we combine it with a self-awareness phenomenon to mitigate this issue. By self-awareness here, we refer to the solution derived from the image itself in the form of saliency cue, which can also be unreliable if applied alone. Nevertheless, combining these two paradigms together can lead to a better co-localization ability. Specifically, we introduce a dynamic mediator that adaptively strikes a proper balance between the two static solutions to provide an optimal solution. Therefore, we call this method \textit{ASOC}: Adaptive Self-aware Object Co-localization. We perform exhaustive experiments on several benchmark datasets and validate that weak-supervision supplemented with self-awareness has superior performance outperforming several compared competing methods. 
\end{abstract}

\section{Introduction}
    
This paper addresses the problem called object co-localization, where the goal is to simultaneously localize the dominant objects of images in a collection of semantically similar images. By localizing, we mean obtaining a tight bounding-box across such objects. In such an image collection, the commonness principle can be very well exploited to perform effective localization of the objects. Such inclusion of the semantically similar images together also provides a mechanism of weak-supervision that can be leveraged in discovering objects. With the same intent, several effective saliency detection methods have also been developed recently with the rise of deep learning techniques. Such saliency detection can be regarded as a self-awareness phenomenon. However, most works on object co-localization either neglect the self-awareness phenomenon or try to strike a balance between weak-supervision and self-awareness. In this work, we attempt to solve the object co-localization problem while taking the second paradigm, but in an adaptive manner.

Most works balance both the phenomenon, self-awareness and weak-supervision, using a certain tuning parameter. Such balancing parameters are actually very difficult to set because the optimal one can vary case-by-case. Instead of relying upon such a balancing parameter, this paper tries to answer if such balancing can be done adaptively to facilitate the application of joint processing algorithms on any dataset without having to set such a parameter. We motivate our work by considering a potential strategy to introduce another factor, a mediator, which actively tries to solve the conflict of maintaining a trade-off between self-awareness and weak-supervision. 

In terms of challenges, there are hardly any prior works that deal with such a problem on adaptive joint processing. There is only one work \cite{8269367} that peeks into this problem, but it relies on quality factor, which is very subjective in nature. It will be interesting to see if an objective approach can be explored. The next challenge is that the adaptive solution should be robust enough so that a joint processing algorithm can be applied on any dataset without any overhead of parameter tuning. Also, there is not much work done on object co-localization, but it has great potential to facilitate applications such as content-based image retrieval since if we know where the object is located then we can employ matching algorithms in the focused region of interest instead of the entire image.

We propose a mediator based adaptive, self-aware object co-localization method where the solution (a mediator) of the previous iteration itself takes part in the optimization and gets updated iteratively. Initially, the mediator completely supports weak supervision by having the same recommendation as weak supervision, and it may start withdrawing that support if even both of them together (mediator and weak-supervision) are not able to supersede self-awareness in achieving the self-aware object localization goals. The mediator keeps getting updated with the last solution until convergence is reached. Since such a strategy does not involve any parameters except the simple tolerance level for convergence, such an algorithm will be effective across the datasets without the need for parameter setting. 

Our contributions are two-fold: 1) To the best of author's knowledge, it is the first work to study the adaptiveness of balancing self-awareness and weak-supervision objectively in object localization tasks. 2) We demonstrate that the proposed method is able to achieve improved results consistently over state-of-the-art of both self-awareness and weak-supervision based localization methods across the compared benchmark datasets. 

The overall architecture of the paper is as follows. Background about the related works is provided in Section \ref{section:related}. The proposed methodology is explained in Section \ref{sec:methodology}. Detailed experimental results and discussion are provided in Section \ref{sec:results} followed by conclusion in Section \ref{sec:conclusion}.

\section{Related Work}
\label{section:related}

Co-saliency refers to concurrently getting the most salient common object across all the images. Co-saliency detection has been extremely beneficial in various object discovery problems. Chang \textit{et al.} \cite{chang2011co} utilizes the co-saliency cue effectively in the co-segmentation problem \cite{li2018deep,8099896,chen2018semantic,9343773}. There have been previous attempts like in ~\cite{fu2013cluster} to fuse various cues, but they all fuse cues spatially. Other similar fusion approaches~\cite{8269367,7484309} fuse raw saliency maps of different images to generate co-saliency maps. All these techniques fuse spatially. \cite{ACPR2017/JingLou,6675796} propose hierarchy based co-saliency detection. While the hierarchy represented in \cite{ACPR2017/JingLou} depicts different scales of the image. \cite{6675796} employs hierarchical segmentation to obtain the co-saliency.
In \cite{zhang2019co}, authors leverage the hierarchical image properties to refine the coarse co-salient segmentation mask obtained using deep networks to get fine co-saliency maps.


The pioneering work on object co-localization problem was introduced by \cite{6909586}, which tries to handle noisy datasets with the ability to avoid assigning the bounding box if the image does not contain the common object. The performance was further improved in \cite{joulin2014efficient}. Next, the authors in \cite{cho2015unsupervised} take a leap over the constraint of even weak supervision and propose a generic co-localization where objects across the images need not be even common. However, since they still use image collection, it's called unsupervised object co-localization, whereas the earlier one is called weakly-supervised object co-localization. Slightly different from the co-localization, there are some bounding-box propagation algorithms \cite{vezhnevets2014associative} where some images already have bounding boxes and they are utilized to localize the unannotated images. It is similar to a supervised scenario and can be called supervised object co-localization. In \cite{wei2019unsupervised}, authors provide a Deep Descriptor Transforming (DDT) technique where they leverage the use of pre-trained convolutional features and utilize the convolutional activations to act as a detector for finding common objects across pool of unlabeled images, i.e. unsupervised co-localization.

Although some of these methods take both weak-supervision (commonness for unsupervised case) and self-awareness into consideration, they all try to balance between the two paradigms. In contrast, the proposed method proposes how to automate this balancing task. The authors in \cite{8269367} attempt to solve it in an automated way, but they take a subjective approach to quality estimation. However, we take an entirely objective approach, which will be discussed in the subsequent sections.



\section{Proposed Method}
\label{sec:methodology}
\subsection{Overview}
Given a set of semantically similar images, the task of object co-localization is to jointly localize the objects in such images. Let $\mathcal{I}=\{I_1,I_2,\cdots,I_n\}$ denote set of $n$ images. Similarly, let $\mathcal{S}=\{S_1,S_2,\cdots,S_n\}$ and $\mathcal{C}=\{C_1,C_2,\cdots,C_n\}$ denote corresponding sets of saliency and co-saliency maps of those images, respectively. Note that any off-the-shelf methods can be used for saliency and co-saliency detection. While saliency detection uses single image $I_i$ for self-awareness, the co-saliency detection uses the entire $\mathcal{I}$ for the weak supervision. To obtain an appropriate bounding box, we essentially have 4 unknown values: topmost-row ($t_i$) number, bottommost-row ($b_i$) number, leftmost column ($l_i$) number, and rightmost column ($r_i$) number of the pixels within the bounding box. Let these values be clubbed in a column vector, $z_i=[t_i,b_i,l_i,r_i]^T$ for the $i-th$ image, $I_i$.

\begin{figure*}[hbtp]
\centering
{
\includegraphics[width=0.95\linewidth]{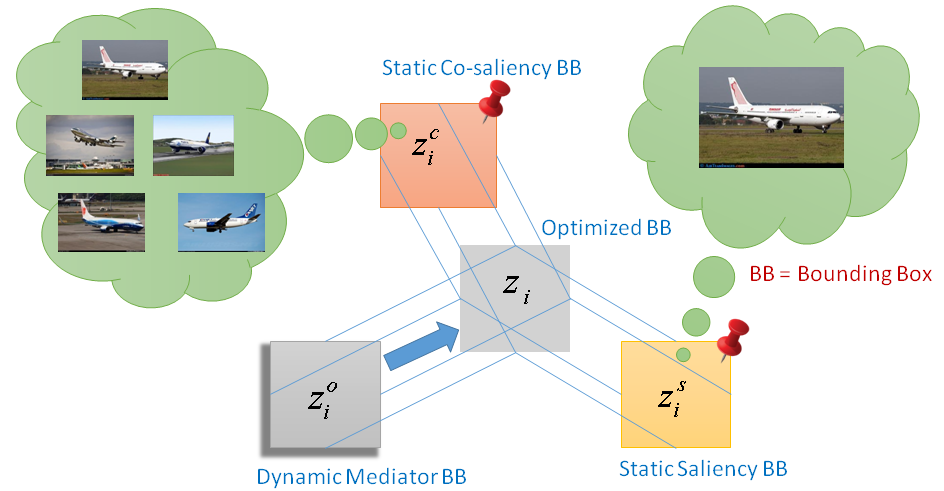}
\caption{Proposed Method: Initially, we have a static saliency map and a static co-saliency map, which yield the static bounding boxes, denoted as $z^c_i$ and $z^s_i$, respectively. In an attempt to strike a proper balance between the two, we introduce what we call a dynamic mediator bounding-box $z^o_i$ (initially, it's equal to $z^c_i$). It acts as the latest bounding box required in our iterative optimization in search for an optimized bounding box $z_i$.}
\label{fig:overview}
}
\end{figure*}

Both saliency and co-saliency can yield bounding boxes, which we call reference bounding boxes. These bounding boxes may not be optimal but can be leveraged. We can obtain these reference bounding boxes by simply thresholding using thechniques like Otsu’s thresholding $\phi(\cdot)$, and finding the extreme pixels in the four directions. Let $z_i^s \text{ and } z_i^c$ be such bounding boxes derived from the saliency map and co-saliency map, respectively. Note that the subscripts of `s’ and `c’ here is to indicate whether they are derived from saliency or co-saliency. The same subscripts will go for the constituents of these bounding box vectors as well, which means $z^c_i=\{t^c_i,b^c_i,l^c_i,r^c_i\}$. Our goal is to find an optimal $z_i$ that is self-aware (i.e., it respects saliency’s $z_i^s$) while complying with $z_i^c$ formed by the co-saliency. As it was motivated earlier, to perform self-aware object co-localization, we take an iterative approach where we take the previous state of the required bounding box into account to update it iteratively. Since $z_i$ is considered the bounding box to be determined, we denote $z^o_i$ as the old bounding-box of the last iteration. Note that $z^o_i$ is initialized as $z^c_i$ at the beginning. Such an arrangement ensures that there is a bias towards co-saliency initially, and this bias, however, is affected only when self-aware co-localization goals are not met. Fig. \ref{fig:overview} shows the workflow of the proposed method and how an optimal $z_i$ is obtained at any iteration with the three reference bounding boxes available.

\subsection{Objective Function}
Given $z^c_i$, $z^s_i$, and $z^o_i$ for an image $I_i$ at any iteration, our goal is to find the optimal $z_i$ that balances between both weak-supervision ($z^c_i$) and self-awareness ($z^s_i$). Additionally, for a smooth transition, $z_i$ shouldn't abruptly change from the old one, i.e., $z^o_i$. Our objective function that keeps all of these into account can be written as follows:

\begin{equation}\label{eq:of}
\begin{split}
\min &  \sum\limits_{k\in\{c,s,o\}}(z_i-z_i^k)^T M^k_i(z_i-z_i^k)\\
    s.t. \text{            } &
b_i>t_i,t_i>=min(\mathcal{E}^1_i), b_i<=max(\mathcal{E}^1_i), \\
& r_i>l_i, l_i>=min(\mathcal{E}^2_i),r_i<=max(\mathcal{E}^2_i),\\
\end{split}
\end{equation}
where we basically try to minimize the collective costs of deviating from the three reference bounding boxes ($z_i^c,z_i^s,\text{ and } z_i^o$) while finding the optimal $z_i$. However, there are certain constraints that need to be taken into consideration while obtaining such an optimal bounding-box: (1) The topmost-row number ($t_i$) should be lower than bottommost-row number ($b_i$), and they should be within minimum and maximum values present in $\mathcal{E}_1$, which is a set of row-numbers of all the edge pixels in the image. (2) The same constraints follow for the horizontal direction (involving $l_i$ and $r_i$), where, similar to $\mathcal{E}_1$, $\mathcal{E}_2$ comprises of column-numbers of all the edge pixels in the image. 

The terms are designed such that if $z_i$ doesn't deviate from a particular $z^k_i$, the cost of deviation from that particular reference bounding box will become zero. Moreover, the deviations have been appropriately weighted by the different rejection cost matrices ($M_i^c, M_i^s, and \text{ } M_i^o$), which will be discussed in the next section, according to the importance of the concerned reference bounding box in accomplishing self-aware object co-localization goals.      

\subsection{Rejection Cost Matrices}
In order to determine how important a particular reference bounding box is, in achieving self-aware object co-localization, we need an absolute object prior, say $A_i$, which accounts for both weak supervision and self-awareness in a united manner, as computed below:
\begin{equation}
    A_i(p)=\begin{cases}
    \frac{q^s_i*S_i(p)+q^c_i*C_i(p)}{q^s_i+q^c_i}, if \text{ }S_i(p)<\phi(S_i) \& C_i(p)>\phi(C_i)\\
    S_i(p), else
    \end{cases}
\end{equation}
where the $A_i$ value of a pixel $p$, i.e. $A_i(p)$, is more or less like that in $S_i$ unless its $S_i$ value is lower and its $C_i$ value is higher than their respective Otsu's threshold ($\phi(\cdot)$) values. For such pixels, we assign the weighted average value from both the maps, with weights $q^s_i$ and $q^c_i$ being quality scores of the two maps. The idea here is to enhance the saliency value of a pixel if it has a good co-saliency. We compute these scores using \cite{7457899}. However, we refrain from using this $A_i$ entirely in figuring out the importance of reference bounding box. Recall that bounding box $B^o_i$ (formed by $z^o_i$) is initialized as $B^c_i$ (formed by $z^c_i$), stating that we rely upon $B^c_i$ more. In that case if we use the entire one, $B^o_i$ is quite likely to be dislodged if there is a spurious salient region elsewhere in the image. Such dislodging should happen only when $B^c_i$ is not good as compared to the corresponding $B^s_i$ (formed by $z^s_i$). Therefore, only those regions of $A_i$ will be considered at any point that overlaps with $B^o_i$. This ensures that $B^o_i$ will not get dislodged unnecessarily. Let $B^a_i$ be the tight bounding box around those overlapping regions of $A_i$, with the required constituents being denoted as $\{t^a_i,b^a_i,l^a_i,r^a_i\}$. In order to derive out the importance of each reference bounding box, we compare each of them with $B^a_i$. Now any $M^k_i$ cost matrix (in \eqref{eq:of}) is defined as follows:
\begin{equation}\label{eq:cost}
    M^k_i=-J(B^a_i,B^k_i)*log(\rho(B^a_i,B^k_i))
\end{equation}
where we multiply the Jaccard Similarity (J) score of the two bounding boxes with their spatial deviation cost matrix, where $\rho(B^a_i,B^k_i)$ is a matrix as described below:  
\begin{equation}
    \rho(B^a_i,B^k_i)=\begin{bmatrix}
\frac{|t^a_i-t^k_i|}{\emph{height}}& 1 & 1 & 1\\
1 & \frac{|b^a_i-b^k_i|}{\emph{height}}&  1 & 1\\
1 & 1 & \frac{|l^a_i-l^k_i|}{\emph{width}} &   1\\
1 & 1 & 1 & \frac{|r^a_i-r^k_i|}{\emph{width}}\\
\end{bmatrix}
\end{equation}
The idea is to have high costs for the bounding boxes having a good overlap with $B^a_i$ and similar coordinates as $B^a_i$. Note that the operations (multiplication and log) in Eqn.~\eqref{eq:cost} are element-wise operations. Note that `\emph{height}' and `\emph{width}' denote image dimensions.  


\begin{table*}[]
\centering
\caption{Comparisons of the CorLoc metric with state-of-the-art co-localization methods on VOC 2007. `-' indicates that the authors have not provided those results in the respective paper.}
\begin{tabular}{llllllllllll}
\hline\hline
Method        & aero & bike  & bird  & boat   & bottle & bus   & car  & cat   & chair & cow  & table \\ \hline\hline
Joulin et al.\cite{joulin2014efficient} & 32.8 & 17.3  & 20.9  & 18.2   & 4.5    & 26.9  & 32.7 & 41.0  & 5.8   & 29.1 & \textbf{34.5}  \\ 
SCDA  \cite{wei2017selective}        & 54.4 & 27.2  & 43.4  & 13.5   & 2.8    & 39.3  & 44.5 & 48.0  & 6.2   & 32.0 & 16.3  \\ 
Cho et al. \cite{cho2015unsupervised}   & 50.3 & 42.8  & 30.0  & 18.5   & 4.0    & 62.3  & \textbf{64.5} & 42.5  & 8.6   & 49.0 & 12.2  \\ 
Li et al.  \cite{li2016image}   & 73.1 & 45.0  & 43.4  & 27.7   & 6.8    & 53.3  & 58.3 & 45.0  & 6.2   & 48.0 & 14.3  \\ 
Vora et al.\cite{vora2018iterative}  &- & -  & - & -   & -    & -  & -& -  & -   & - & -  \\
Vo et al.\cite{vo2020toward}  &- & -  & - & -   & -    & -  & -& -  & -   & - & -  \\
DDT   \cite{wei2019unsupervised}        & 67.3 & 63.3  & 61.3  & 22.7   & 8.5    & \textbf{64.8}  & 57.0 & \textbf{80.5}  & 9.4   & 49.0 & 22.5  \\
DDT+  \cite{wei2019unsupervised}        & 71.4 & \textbf{65.6}  & 64.6  & 25.5   & 8.5    & \textbf{64.8}  & 61.3 & \textbf{80.5}  & 10.3  & 49.0 & 26.5  \\\hline
Ours ASOC          &\textbf{78.6}  & 42.4  &\textbf{72.4}   & \textbf{50.3}   &  \textbf{13.1}  & 58.1  &64.4  &77.4   & \textbf{18.9}  &\textbf{76.6}  & 18.5 \\ \hline \hline
Method        & dog  & horse & mbike & person & plant  & sheep & sofa & train & tv    & Mean &       \\ \hline\hline
Joulin et al.\cite{joulin2014efficient}  & 31.6 & 26.1  & 40.4  & 17.9   & 11.8   & 25.0  & 27.5 & 35.6  & 12.1  & 24.6 &       \\ 
SCDA  \cite{wei2017selective}        & 49.8 & 51.5  & 49.7  & 7.7    & 6.1    & 22.1  & 22.6 & 46.4  & 6.1   & 29.5 &       \\
Cho et al. \cite{cho2015unsupervised}   & 44.0 & 64.1  & 57.2  & 15.3   & 9.4    & 30.9  & 34.0 & 61.6  & 31.5  & 36.6 &       \\
Li et al.  \cite{li2016image}   & 47.3 & 69.4  & 66.8  & 24.3   & 12.8   & 51.5  & 25.5 & 65.2  & 16.8  & 40.0 &       \\
Vora et al.\cite{vora2018iterative}  &- & -  & - & -   & -    & -  & -& -  & -   & 35.1   \\
Vo et al.\cite{vo2020toward}  &- & -  & - & -   & -    & -  & -& -  & -   & 46.7   \\
DDT \cite{wei2019unsupervised}          & \textbf{72.6} & 73.8  & 69.0  & 7.2    & \textbf{15.0}   & 35.3  & 54.7 & 75.0  & 29.4  & 46.9 &       \\
DDT+  \cite{wei2019unsupervised}        & \textbf{72.6} & 75.2  & 69.0  & 9.9    & 12.2   & 39.7  & \textbf{55.7} & 75.0  & \textbf{32.5}  & 48.5 &  \\\hline
Ours ASOC         & 68.9 & \textbf{78.7}  & \textbf{73.5} & \textbf{54.3}   & 13.1   &\textbf{65.6} &45.0 & \textbf{77.0} & 21.1  & \textbf{53.4}  &  \\
\hline    
\end{tabular}
\label{tab:tabpascal}
\end{table*}
\subsection{Proposed Solution \& Implementation Details}
The Eqn.\eqref{eq:of} can be easily converted to a quadratic programming problem with linear constraints. Such a problem can be solved using the \emph{quadprog()} function available in Matlab. 

We stop our iterative optimization when the L2-norm value $||z_i-z_i^o||^2_2$ between solutions of any two consecutive iterations is $<\epsilon$. Here, we set $\epsilon$ as 2. Usually, our objective function converges within 3-5 iterations. However, if it doesn't converge due to some reason even after 30 iterations, we break the loop forcibly. Such a case arises when there are more than one valid solutions, with each one manifesting at alternate iterations.

\begin{table}[]
\centering
\footnotesize

\caption{Comparisons of CorLoc on OD100 dataset (subset of Internet Images dataset). `-' indicates that the authors have not provided those results in the respective paper.}
\begin{tabular}{l|lll|l}
\hline\hline
Method            & Airplane & Car   & Horse & Mean  \\ \hline\hline
Joulin et al. \cite{joulin2010discriminative}    & 32.93    & 66.29 & 54.84 & 51.35 \\ 
Joulin et al. \cite{joulin2012multi}    & 57.32    & 64.04 & 52.69 & 58.02 \\ 
Rubinstein et al.\cite{rubinstein2013unsupervised} & 74.39    & 87.64 & 63.44 & 75.16 \\ 
Tang et al. \cite{6909586}      & 71.95    & 93.26& 64.52 & 76.58 \\ 
SCDA \cite{wei2017selective}            & 87.80    & 86.52 & 75.37 & 83.20 \\ 
Cho et al.  \cite{cho2015unsupervised}      & 82.93    & 94.38 & 75.27 & 84.19 \\ 
Vora et al.  \cite{vora2018iterative}      & 43.9   & 65.17 & 45.16 & 51.41  \\ 
Vo et al.  \cite{vo2020toward}      & -    & - & - & 90.2  \\ 
DDT\cite{wei2019unsupervised}          &     \textbf{91.46}     & 95.51     &  77.42      &   88.13     \\ 
DDT+\cite{wei2019unsupervised}         &   \textbf{91.46}       &  94.38     &    76.34    &      87.39   \\ 
Our ASOC          &  90.24        &   \textbf{98.88}    &   \textbf{86.02}    &   \textbf{91.71}    \\ \hline
\end{tabular}
\label{tab:tabod}
\end{table}

\section{Experiments Results}
\label{sec:results}

All the experiments are performed in a weakly supervised scenario where the images are categorized as per their classes. We compute the co-saliency for such an image collection and use it to balance the mediating factor benefited from self-awareness and weak-supervision.


\begin{figure*}[hbtp]
\centering
{
\includegraphics[width=0.95\linewidth]{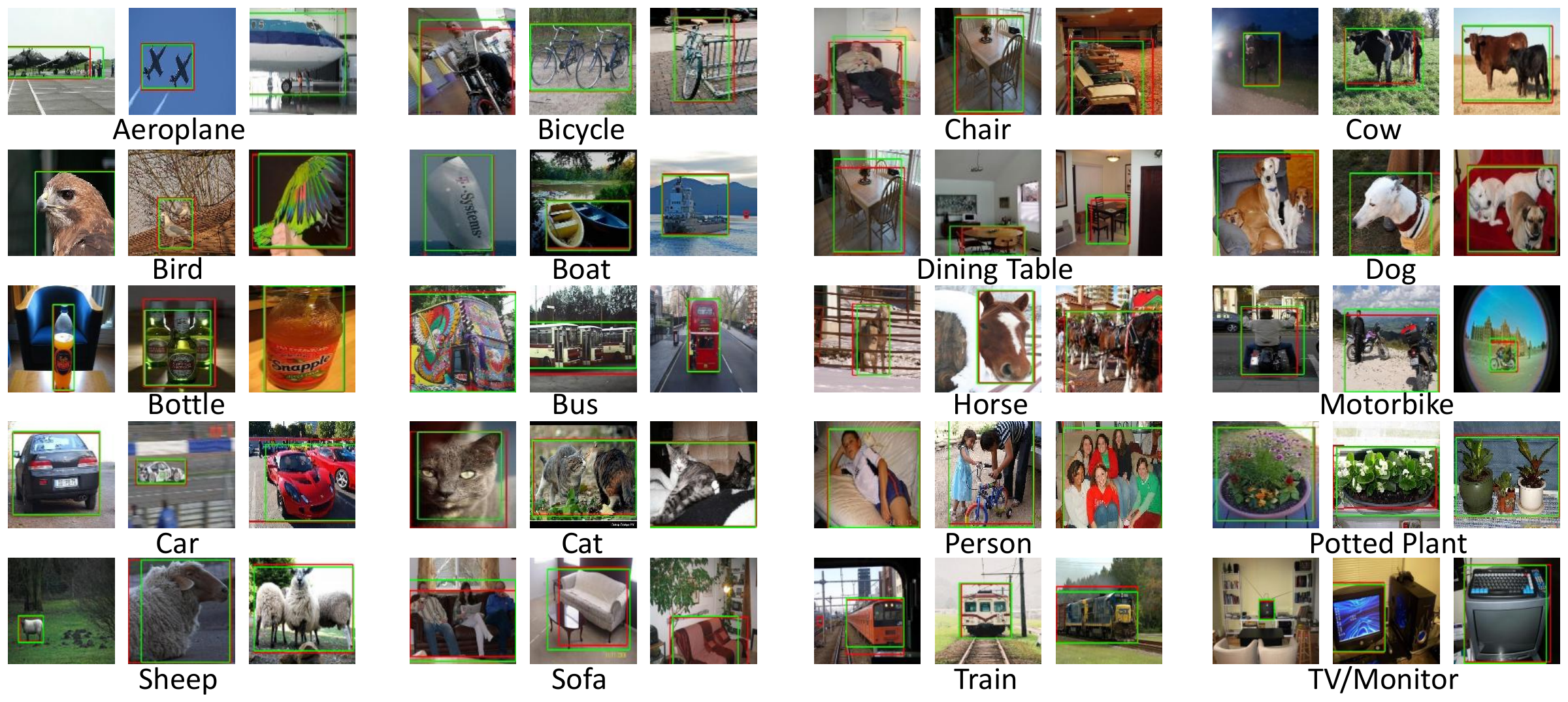}
\caption{Qualitative examples of PASCAL VOC 2007 dataset. The red bounding boxes in these figures are the ground truth boxes and green indicate the colocalization boxes obtained by the proposed method \textit{ASOC}. (Best viewed in color)}
\label{fig:pascalvoc}
}
\end{figure*}

\subsection{Dataset Used and Evaluation Metrics}
We evaluate the proposed method on Internet Images dataset\cite{rubinstein2013unsupervised}, which comprises of three categories: Aeroplane, Car, and Horse. Most of the existing works have reported their results on the same benchmark dataset. We also report our results on PASCAL VOC 2007 dataset\cite{everingham2015pascal}. Internet images dataset has segmentation masks, using which tight bounding-boxes are generated as ground-truth bounding-boxes. In PASCAL VOC 2007 dataset, the ground truth bounding boxes are provided. It consists of 20 classes with a training+validation set (5011 images) and test set (4952 images). In Pascal VOC 2007 dataset, for multiple object instances present in an image, we create a single tight bounding box enclosing all individual grouth-truth boxes for that object class and use it as the ground truth bounding box.

We perform the experiments on PASCAL VOC dataset in congruent lines with \cite{li2016image, cho2015unsupervised, joulin2014efficient} where all images in the trainval set are utilized except for the ones which contain only difficult or truncated object instances. For Internet Images  dataset, we have utilized the subset of 100 images per category as followed in \cite{cho2015unsupervised, rubinstein2013unsupervised} in order to have a fair comparison with other competing methods. We refer to this subset of the Internet images dataset as OD100 in our experiments. 


Existing works on object co-localization widely use the correct localization  (CorLoc) metric for evaluation. The CorLoc metric is defined as the percentage of images that obtain correct localization results according to the criteria IoU (intersection-over-union)$>=0.5$.

\subsection{Results}

In Table \ref{tab:tabpascal}, we compare our work with several existing works on Pascal VOC 2007 dataset. The proposed ASOC method is able to outperform all other methods with a relative gain of at least 10\% in terms of 'Mean' CorLoc score. It performs best in 11/20 categories. Similarly, in Table~\ref{tab:tabod}, we compare our results with other competing methods on OD100 dataset. Our ASOC method outperforms every other method in terms of 'Mean' CorLoc score here as well. We achieve the best results in 2/3 categories.   

\begin{figure}[t]
\centering
{
\includegraphics[width=0.95\linewidth]{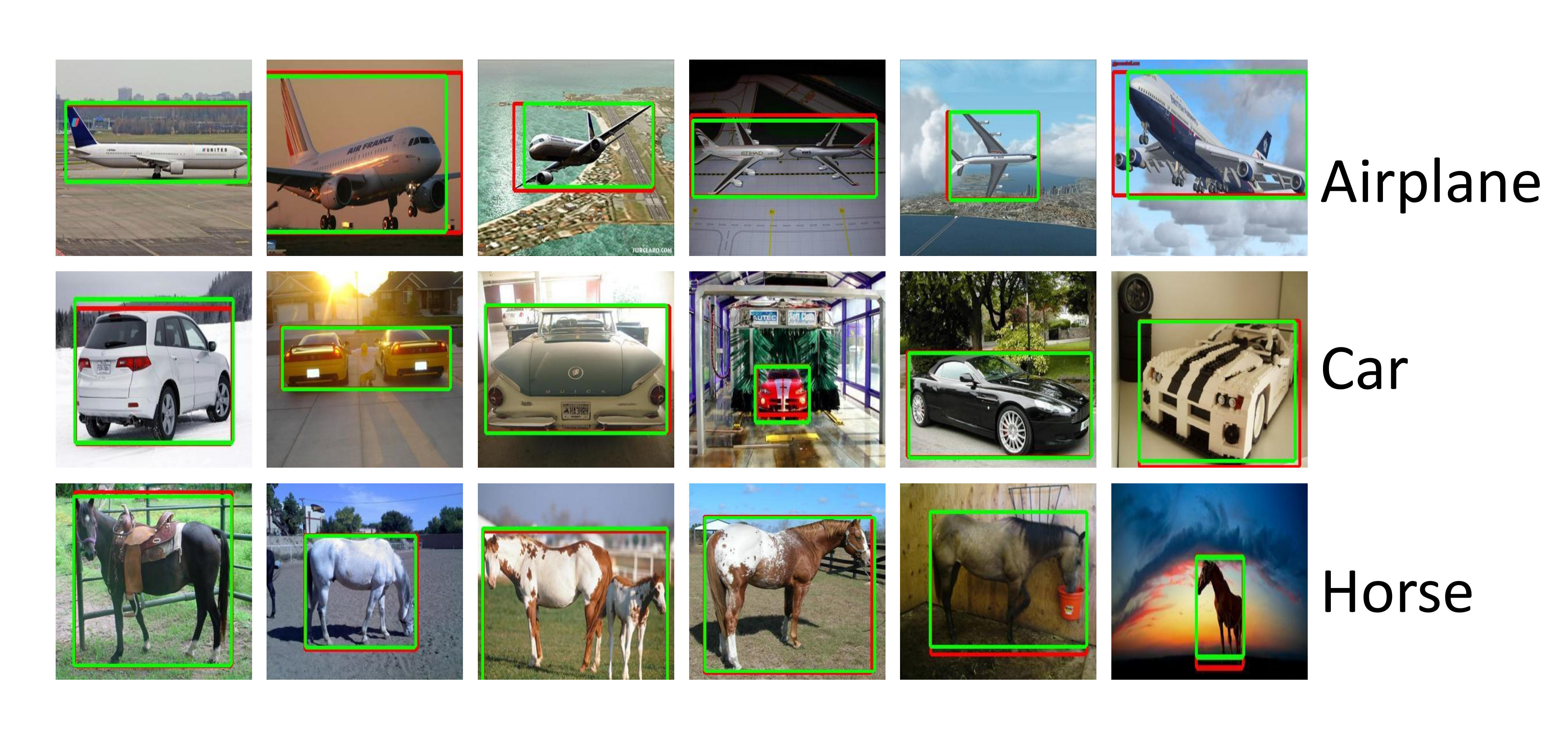}
\caption{Qualitative examples of OD100 dataset. The red bounding boxes in these figures are the ground truth boxes and green indicate the colocalization boxes obtained by the proposed method \textit{ASOC}. (Best viewed in color)}
\label{fig:od}
}
\end{figure}

In Figures \ref{fig:pascalvoc}  \& \ref{fig:od}, we demonstrate the qualitative co-localization results on PASCAL VOC 2007 and OD100 datasets, respectively. We show the variability captured in terms of orientation (e.g. in Pascal VOC 2007 classes: train, bus, motorbike), scale (e.g. in Pascal VOC 2007 classes: aeroplane, bird, dog) and number of object instances (e.g. in Pascal VOC classes: cow, bicycle, sheep, horse, pottedplant etc.). This demonstrates the robustness of the proposed method and its ability to co-localize simultaneously multiple instances of objects and work well in  various challenging scenarios for providing tight bounding boxes.

\section{Conclusion}
\label{sec:conclusion}
In this paper, we have proposed a novel self-aware object co-localization method that leverages both self-awareness (saliency) and weak-supervision (co-saliency) to effectively localize the common objects in a collection of images. We develop an iterative framework where the required bounding box gets updated after every iteration while being part of the optimization. Our results on two publicly available datasets, namely OD100 dataset and VOC 2007 dataset, demonstrate excellent results, surpassing the existing works comfortably in terms of co-localization results in weakly-supervised scenario.

\bibliographystyle{IEEEbib}
\bibliography{icme2021template}

\end{document}